# Local Distance Metric Learning for the Nearest Neighbor Algorithm

H. Rajabzadeh,    M. Z. Jahromi,    M.S. Zare,    S.M. Fakhrahmad

**Department of Computer Science and Engineering, Shiraz University**

rajabzadeh@cse.shirazu.ac.ir    zjahromi@shirazu.ac.ir    mzare@cse.shirazu.ac.ir    fakhrahmad@shirazu.ac.ir

**Abstract-**

Distance metric learning is a successful way to enhance the performance of the nearest neighbor classifier. In most cases, however, the distribution of data does not obey a regular form and may change in different parts of the feature space. Regarding that, this paper proposes a novel local distance metric learning method, namely Local Mahalanobis Distance Learning (LMDL), in order to enhance the performance of the nearest neighbor classifier. LMDL considers the neighborhood influence and learns multiple distance metrics for a reduced set of input samples. The reduced set of samples which are denoted as prototypes try to preserve local discriminative information as much as possible. The proposed LMDL can be kernelized very easily, which is significantly desirable in the case of highly nonlinear data. The quality as well as the efficiency of the proposed method are assessed through a set of different experiments on various datasets. The obtained results show that LDML as well as the kernelized version are superior to the other related state-of-the-art methods.

## 1. Introduction

Learning a proper distance function over the data points plays a crucial role in many domains, specifically in machine learning domain [1]. The ultimate goal of the distance metric learning methods is to keep similar samples close together while putting them far away from dissimilar points. Since the decision making process in nearest neighbor (NN) algorithm is based on the way that distances are computed, learning a proper distance metric can significantly improve the performance of this popular classification algorithm. Besides, distance metric learning plays a vital role in many machine learning tasks [2][3]. In this way, learning a Mahalanobis distance metric, which is a global distance metric, gets lots of attention due to its efficiency and simplicity in solving complicated problems [2][3]. In the case of multimodal data distributions, however, such a global metric does not work properly and cannot satisfy all the constraints on a global point of view. Capturing the local information between data points have been considered as a solution to cope with this deficiency. In this way, the first group of methods tries to learn multiple distance metrics while the second group tries to learn one metric by imposing some local constraints in order to capture local information. Having a huge number of constraints, which is not unusual in today's real-world tasks, the second group of methods may face many obstacles in their optimization processes leading them to employ many relaxations in order to ease the problem. On the other hand, the risk of overfitting can be considered as a main drawback of the first group. Besides, most of them employ an objective function which are not closely related to the notion of NN decision rule at all.

Having said this, this paper proposes a supervised distance metric learning method avoiding all of the aforementioned shortages. At the first place, LMDL selects a small set of samples, namely prototypes, and then learns a Mahalanobis distance metric for each prototype based on a closely related objective function proposed in ref. [4] and attracted a lot of interest [5][6][7][8][9]. Furthermore, the proposed learning procedure adjusts the prototype positions in order to minimize the objective function as well. Selecting prototypes mitigates the risk of overfitting while preserving the notion of locality. Additionally, we develop the kernelized version of the proposed method, namely kLMDL, which is very desirable in the case of highly nonlinear data structures. LMDL, also, can be considered as an extension of LPD (learning prototype and distance) [7]. While LPD learns a weight vector for each prototype, LMDL learns a Mahalanobis distance metric for each prototype. Moreover, our method is able to be used in kernel space as well. These changes turn our method into a very powerful one in dealing with real-world problems.

We performed a number of different evaluations and the experimental results clearly demonstrate that LMDL as well as kLMDL significantly increases the predictive performance over the other related state-of-the-art metric learning methods. In a nutshell, the major contribution of the paper can be summarized as follows:

- Learning exactly one Mahalanobis metric for each prototype based on a closely related objective function to the NN decision rule
- Adjusting the prototype position
- Developing the kernelized version of the proposed method based on the concept of "kernel trick" making it more efficient and highly flexible.

The rest of the paper is organized as follows. Section 2 provides a brief history of some related distance metric learning methods. Section 3 presents LMDL as well as kLMDL in details. Section 4 presents and discusses the experimental results. Finally, Section 5 concludes the paper.

## 2. Related work

This section summarizes a brief history of supervised distance metric learning methods which are related to our work. The work proposed by Xing et al. [10], apparently, can be considered as the first attempt to learn a Mahalanobis distance metric which uses a positive semi-definite (PSD) formulation to maximize the sum of distances between different-class samples while keeping the same-class samples close together. Neighborhood component analysis (NCA) [11] uses a relative, but non-convex, objective function to the NN decision rule trying to minimize an estimation of the expected leave-one-out (LOO) error of the NN in the projection space. In the same way, MCML (maximally collapsing metric learning) [12] uses a convex formulation in order to collapse the same-class samples to a single point. Although MCML benefits from a convex formulation, it suffers from high computational complexity. One of the most popular Mahalanobis distance metric learning methods is LMNN (large margin nearest neighbor) [13][14] considering the local information by imposing constraints in a local manner. LMNN and those proposed in refs. [15][16][17][18] learn a global metric based on the semi-definite programing and, accordingly, suffer from high computational complexity. In the following, LMCA (large margin component analysis) [19] tries to mitigate the computational complexity of the LMNN by solving for a $d$ by $k$ matrix instead of a $d$ by $d$ one where $d$ is the dimensionality of the input space and $k \ll d$. Recently, a global metric learning method is proposed in ref. [20] trying to maximize the Jeffery divergence between two multivariate Gaussian distributions derived from local pairwise constraints. However, it fails to take into account all of the local information when the number of constraints rises up to a considerable level. More recently, Harandi et al. [21] have proposed a framework that jointly learns a lower dimensional mapping space and a global distance metric.

Parametric Local Metric Learning (PLML) [22] learns a local metric per each point based on a linear combination of some predefined basis metrics. However, PLML depends on a manifold assumption and suffers from a large number of parameters. In contrast to the PLML, generative local metric learning (GLML) [23] learns an independent metric per each point through a generative process and minimizes the NN expected error under some assumptions for the class distributions. In the following, sparse compositional metric learning (SCML) [24] uses rank-one matrices to construct one metric per each point and, therefore, reduces the number of parameters compared to the one in [22]. LDDM (local discriminative distance metrics) [25] is a local metric learning method learning multiple distance metrics, one metric per each exemplar point, with the aim of optimizing local compactness as well as local discrimination. Similar to the above mentioned local methods, however, LDDM suffers from an excessive number of parameters putting it in the risk of overfitting. More recently, Li et al. have proposed [26] a metric learning method based on the eigenvalue decomposition that can be used for both global and local views. Although MLEV is faster than some other methods, it enforces orthogonality constraints on the learned metric and, thus, performing dimensionality reduction, and not metric learning at all.

In contrast to the aforementioned methods, our proposed technique concurrently benefit from the following advantages making it a powerful metric for the kNN classification tasks and inspiring the idea of our work. First, we use an objective function closely related to the NN decision rule. Second, we use the notion of prototype and learn one local Mahalanobis metric for each prototype in order to capture local discriminative information while preventing the risk of overfitting. Third, we iteratively adjust the position of prototypes to find the best position for them. Finally, we propose the kernelized version of our method which is desirable for highly nonlinear datasets.

3. **Proposed method**

Given a set of $M$ training points $\mathcal{X} = \{(x^1, y^1), \ldots, (x^M, y^M)\}$, where $x^m \in \mathbb{R}^{d \times 1}$ and $y^m \in \{1, 2, \ldots, K\}$ defines the corresponding class label, the ultimate goal is to learn a set of Mahalanobis metrics $\mathbb{W} = \{W\}_{s=1}^{S}$ where $W^s \in \mathbb{R}_+^{d \times d}$ is a positive semi-definite matrix and corresponds to the $s^{th}$ member of a set of randomly selected prototypes $\mathbb{P} = \{(p^1, y^1), \ldots, (p^S, y^S)\}$, $p^s \in \mathbb{R}^{d \times 1}$ and $S \ll M$. In this setting, vectors and matrices are respectively denoted by boldface lowercase and boldface uppercase letters. Accordingly, the scalars are denoted by uppercase or lowercase letters. Also, to have a compact representation of the parameters, suppose that $\mathcal{W} \in \mathbb{R}^{(d \times d) \times S}$ is a matrix in which the $s^{th}$ column of $\mathcal{W}$ represents the vectorized form of $W^s$. Similarly, $\mathcal{P} \in \mathbb{R}^{d \times S}$ is a matrix in which $s^{th}$ column of $\mathcal{P}$ holds $p^s$ and the $i^{th}$ column of $X \in \mathbb{R}^{d \times M}$ is the $i^{th}$ point in set $\mathcal{X}$.

Using the above notations, the squared Mahalanobis distance between $s^{th}$ prototype and $i^{th}$ point in the input space is given by:

$$d_{W^s}^2(x^i, p^s) = \|x^i - p^s\|_{W^s}^2 = (x^i - p^s)^T W^s (x^i - p^s) \tag{1}$$

where $W^s \in \mathbb{R}_+^{d \times d}$ is a symmetric PSD matrix defined on the $s^{th}$ prototype. In order to minimize the error rate of the NN algorithm, we use the following objective function which is a close approximation of the NN's error rate and has been proposed in [4].

$$J(\mathcal{W}, \mathcal{P}) = \frac{1}{M} \sum_{x^i \in \mathcal{X}} \mathbb{S}_\beta(R(x^i)), \tag{2}$$

$$R(x^i) = \frac{d_{W^=}^2(x^i, \mathcal{P}^=)}{d_{W^{\neq}}^2(x^i, \mathcal{P}^{\neq})}$$

In Eq. (2), $\mathbb{S}_\beta(z) = \frac{1}{1 - e^{\beta(1-z)}}$ is a sigmoid function and $\mathcal{P}^=$, $\mathcal{P}^{\neq} \in \mathbb{P}$ are, respectively, the nearest same-class and the nearest different-class prototypes of $x^i$, i.e.

$$\mathcal{P}^= = \underset{\substack{p \in \mathcal{P} \\ \text{class}(p) = \text{class}(x)}}{\arg\min} d_{W^=}^2(x, p) \quad , \quad \mathcal{P}^{\neq} = \underset{\substack{p \in \mathcal{P} \\ \text{class}(p) \neq \text{class}(x)}}{\arg\min} d_{W^{\neq}}^2(x, p) \tag{3}$$

Accordingly, $W^=$, $W^{\neq} \in \mathbb{W}$ respectively are the corresponding Mahalanobis metrics of $\mathcal{P}^=$ and $\mathcal{P}^{\neq}$. The parameter $\beta$ defines the slope of sigmoid function and if $\beta$ is large, $\mathbb{S}_\beta(z)$ behaves like the step function more and more.

Based on Eq. (2), the optimization problem can be written as follows:

$$(\mathcal{W}^*, \mathcal{P}^*) = \underset{\substack{\mathcal{W} \in \mathbb{R}^{(d \times d) \times S} \\ \mathcal{P} \in \mathbb{R}^{d \times S}}}{\arg\min} J(\mathcal{W}, \mathcal{P}) \tag{4}$$

$$\text{subject to: } W \geq 0, \forall W \in \mathbb{W}$$

Eq. (4), however, is a semi-definite programing with a non-convex objective function. Using the fact that $W \in \mathbb{W}$ is a symmetric PSD matrix, it can be factorized as $W = \widetilde{W}\widetilde{W}^T$ where $\widetilde{W} \in \mathbb{R}^{d \times p}$ and $p \leq d$. Hence, Eq. (4) changes as:

$$(\widetilde{\mathcal{W}}^*, \mathcal{P}^*) = \underset{\substack{\widetilde{\mathcal{W}} \in \mathbb{R}^{(d \times p) \times S} \\ \mathcal{P} \in \mathbb{R}^{d \times S}}}{\arg \min} J(\widetilde{\mathcal{W}}, \mathcal{P}) \tag{4}$$

where $s^{th}$ column of $\widetilde{\mathcal{W}} \in \mathbb{R}^{(d \times p) \times S}$ is the vectorized form of the matrix $\widetilde{W}^s \in \mathbb{R}^{d \times p}$. In order to minimize $J(\widetilde{\mathcal{W}}, \mathcal{P})$, we propose an iterative gradient-based procedure which needs the following derivatives:

$$\nabla_{\widetilde{W}^=} J(\widetilde{\mathcal{W}}, \mathcal{P}) = \frac{1}{M} \sum_{x^i \in X} \mathbb{S}'_\beta(R(x^i)) \odot \nabla_{\widetilde{W}^=} R(x^i) \tag{5}$$

$$\nabla_{\widetilde{W}^=} R(x^i) = \frac{1}{d^2_{\widetilde{W}^{\neq}}(x^i, \mathcal{P}^{\neq})} \nabla_{\widetilde{W}^=} d^2_{\widetilde{W}^=}(x^i, \mathcal{P}^=)$$

$$\nabla_{\widetilde{W}^=} d^2_{\widetilde{W}^=}(x^i, \mathcal{P}^=) = \nabla_{\widetilde{W}^=} Trace\left(\widetilde{W}^{=^T}(x^i - \mathcal{P}^=)(x^i - \mathcal{P}^=)^T \widetilde{W}^=\right) = 2(x^i - \mathcal{P}^=)(x^i - \mathcal{P}^=)^T \widetilde{W}^=$$

$$\nabla_{\mathcal{P}^=} J(\widetilde{\mathcal{W}}, \mathcal{P}) = \frac{1}{M} \sum_{x^i \in X} \mathbb{S}'_\beta(R(x^i)) \odot \nabla_{\mathcal{P}^=} R(x^i) \tag{6}$$

$$\nabla_{\mathcal{P}^=} R(x^i) = \frac{1}{d^2_{\widetilde{W}^{\neq}}(x^i, \mathcal{P}^{\neq})} \nabla_{\mathcal{P}^=} d^2_{\widetilde{W}^=}(x^i, \mathcal{P}^=)$$

$$\nabla_{\mathcal{P}^=} d^2_{\widetilde{W}^=}(x^i, \mathcal{P}^=) = -2\widetilde{W}^= \widetilde{W}^{=^T}(x^i - \mathcal{P}^=)$$

where $\mathbb{S}'_\beta(z) = \beta \mathbb{S}_\beta(z)(1 - \mathbb{S}_\beta)$ is the derivative of $\mathbb{S}_\beta(z)$ and the notation $\odot$ represents entrywise multiplication. The calculation of $\nabla_{\widetilde{W}^{\neq}} J(\widetilde{\mathcal{W}}, \mathcal{P})$ as well as $\nabla_{\mathcal{P}^{\neq}} J(\widetilde{\mathcal{W}}, \mathcal{P})$ is similar to Eq. (5) and Eq. (6) respectively. In order to update parameters, we use Adadelta rule [27], which is an extension of Adagrad rule and requires no manual tuning of the learning rate. Furthermore, it appears robust to noisy gradient information, different model architecture choices, various data modalities and selection of hyperparameters.

Based on the above derivatives and similar to the learning procedure proposed in [5], Algorithm 1 summarizes the proposed learning algorithm which is an iterative gradient based procedure. As the Algorithm 1 shows, in each iteration, the algorithm visits $x \in X$ and updates those two metrics that have the highest impact in prediction of sample $x$ and represent by $\widetilde{W}^=$ and $\widetilde{W}^{\neq}$. In other words, the algorithm updates $\widetilde{W}^=$ in a way that the sample $x$ gets closer to its nearest same-class prototype, i.e. $\mathcal{P}^=$, while updating $\widetilde{W}^{\neq}$ in a way that it gets far from its nearest different-class prototype, i.e. $\mathcal{P}^{\neq}$. Concurrently, the same-class and different-class nearest prototypes are modified in a way that $\mathcal{P}^=$ moves toward $x$, while $\mathcal{P}^{\neq}$ gets away from $x$.

The proposed LMDL has several fascinating advantages. First of all, setting $p \ll d$, the algorithm needs much smaller number of parameters due to the low rank representation of $\widetilde{\mathcal{W}}$ and gives us an opportunity for scalable problems. Second, as it has been mentioned in [5], only those training samples which are close to the decision boundary contribute to the update of parameters. This stems from the fact that the update steps are weighted by the distance ratio $R(x)$ and windowed by the $\mathbb{S}'_\beta(R(x))$. This means that, the learning algorithm is not sensitive to outliers whose $R(x)$ is very large, since $\mathbb{S}'_\beta(R(x))$ becomes too small and mitigates their effects in learning. On the other hand, the windowing effect prevents learning from those correctly classified samples that are far from decision boundary, i.e. $\{x|u_x \ll 1, x \in X\}$. Finally, not only does the objective function directly related to the final classifier to use, i.e. kNN, but also it relates to the concept of margin maximization which is a very desirable property. For more information, we refer the interested readers to [9].

The number of selected prototypes in each class will be effective on the generalization of LMDL as well as kLMDL. If it is small, the true nature of class distribution may not be captured. On the other hand, if it is large, the generalization power will reduce and the problem of overfitting would be inevitable. Although our proposed optimization problem is not a convex one, the metrics learned by our solution, as the experimental results confirm, are consistently comparable to those computed by globally-optimal methods.

**Algorithm 1 (LMDL)**

//Input: $\mathcal{X}$, $S$: number of prototypes, $\beta$: slope of sigmoid, $\varepsilon$: small constant

//Output: $\widetilde{\mathcal{W}}, \mathcal{P}$

1. Intitialize $\mathcal{W}$ & $\mathcal{P}$ randomly
2. Set $\widetilde{\mathcal{W}}^{new} = \widetilde{\mathcal{W}}$, $\mathcal{P}^{new} = \mathcal{P}$, $\lambda' = \infty$, $\lambda = J(\widetilde{\mathcal{W}}, \mathcal{P})$
3. **while** $(|\lambda' - \lambda| > \varepsilon)\{$
   3.1. $\lambda' = \lambda$
   3.2. **For** $x \in \mathcal{X}$
      3.2.1. $\mathcal{P}^= = findNNSameClass(x, \mathcal{P})$
      3.2.2. $\mathcal{P}^{\neq} = findNNDiffClass(x, \mathcal{P})$
      3.2.3. $R(x) = d^2_{W^=}(x, \mathcal{P}^=)/d^2_{W^{\neq}}(x, \mathcal{P}^{\neq})$
      3.2.4. Calculate $\nabla_{\widetilde{W}^=} J(\widetilde{\mathcal{W}}, \mathcal{P}), \nabla_{\widetilde{W}^{\neq}} J(\widetilde{\mathcal{W}}, \mathcal{P}), \nabla_{\mathcal{P}^=} J(\widetilde{\mathcal{W}}, \mathcal{P}), \nabla_{\mathcal{P}^{\neq}} J(\widetilde{\mathcal{W}}, \mathcal{P})$
      3.2.5. $[\alpha_{\widetilde{W}^=}, \alpha_{\widetilde{W}^{\neq}}, \alpha_{\mathcal{P}^=}, \alpha_{\mathcal{P}^{\neq}}]$ =Use the Adadelta rule to find corresponding learning rates [27]
      3.2.6. $\widetilde{\mathcal{W}}^{=,new} = \widetilde{\mathcal{W}}^= - \alpha_{\widetilde{W}^=} \odot \nabla_{\widetilde{W}^=} J(\widetilde{\mathcal{W}}, \mathcal{P})$
      3.2.7. $\widetilde{\mathcal{W}}^{\neq,new} = \widetilde{\mathcal{W}}^{\neq} - \alpha_{\widetilde{W}^{\neq}} \odot \nabla_{\widetilde{W}^{\neq}} J(\widetilde{\mathcal{W}}, \mathcal{P})$
      3.2.8. $\mathcal{P}^{=,new} = \mathcal{P}^= - \alpha_{\mathcal{P}^=} \odot \nabla_{\mathcal{P}^=} J(\widetilde{\mathcal{W}}, \mathcal{P})$
      3.2.9. $\mathcal{P}^{\neq,new} = \mathcal{P}^{\neq} - \alpha_{\mathcal{P}^{\neq}} \odot \nabla_{\mathcal{P}^{\neq}} J(\widetilde{\mathcal{W}}, \mathcal{P})$
   3.3. $\widetilde{\mathcal{W}} = \widetilde{\mathcal{W}}^{new}$ & $\mathcal{P}^{new} = \mathcal{P}$
   3.4. $\lambda = J(\widetilde{\mathcal{W}}, \mathcal{P})$

Algorithm 1. The proposed learning algorithm

Although the linear metric learning techniques are applied properly in many cases, they are insufficient in the case of nonlinear data, i.e. data with complex and nonlinear decision boundary. The kernel methods are a suitable choice in such cases. The idea of kernel space is to implicitly project original data into a higher dimensional space through a kernel function and learn distance metric in that space. This implicit projection makes it very probable to linearly separate the data in that space. The following paragraphs illustrate the kernelized version of our proposed method, namely kLMDL, making it a powerful metric learning method to deal with highly nonlinear structures.

We start using the fact that the column space of $\widetilde{W}$ is a subset of the column space of $X$, i.e. $\mathcal{C}(\widetilde{W}) \subseteq \mathcal{C}(X)$. Hence, it is possible to reconstruct $\widetilde{W}$ with a linear combination of the columns of $X$, i.e. $\widetilde{W} = XB$ where $B \in \mathbb{R}^{M \times p}$. Replacing $\widetilde{W}$ with $XB$, we can rewrite the Eq. (1) as follows:

$$d_{W^s}^2(x^i, p^s) = \tag{7}$$

$$(x^i - p^s)^T X B^s B^{s^T} X^T (x^i - p^s) =$$

$$(X^T x^i - X^T p^s)^T B^s B^{s^T} (X^T x^i - X^T p^s) =$$

$$(K_i - K_s)^T B^s B^{s^T} (K_i - K_s)$$

where $K_i = X^T x^i$ is the $i^{th}$ column of the matrix $K = [X\ \mathcal{P}]^T [X\ \mathcal{P}] \in \mathbb{R}^{M \times M}$ and $B^s \in \mathbb{R}^{M \times p}$ is a matrix defined on the $s^{th}$ prototype. Eq. (7) vividly shows that the distance function is represented in terms of the inner products between points. As an immediate results of Eq. (7), we can replace the inner product via any suitable kernel function $K$, e.g. Gaussian kernel [28]. In this setting, we need to learn $B^s$ instead of $\widetilde{W}^s$. The required derivatives and the learning procedure will proceed as same as those in the non-kernelized version. It is worthy to note that the size of $M$ can be very large and, therefore, make it hard to calculate the matrix $K$. However, we can always use the Nyström method [29] to deal with large scale optimization problems.

## 4. Experimental results

This section evaluates the performance of LMDL as well as kLMDL and compares them with a number of related state-of-the-art metric learning methods on both artificial and real-world datasets. In all of the experiments reported here, the parameter $\beta$ and the number of prototypes per each class are selected empirically. However, in most of the cases, the parameters $\beta$ and the number of prototypes per each class were set to 10 and 5 respectively. Besides, the RBF kernel has been used as the data kernel and the kernel parameter, i.e. standard deviation, is tuned by 10-fold cross validation on a set values $\{2^{-15}, \ldots, 2^3\}$. The rest of the parameters were initialized randomly and the NN classifier is used as the final classifier for classification.

The first experiment visually explores the discrimination power of the LMDL. Consider three artificial datasets, two classes, eight classes, and five classes, respectively, depicted in Figures 1.a, 2.c, and 2.e (the third one is the well-known Helix dataset). We set two prototypes for each class and use the Algorithm 1 so as to learn LMDL's parameters. To visualize the results, we use the fact that learning a Mahalanobis distance metric is equivalent to learn a projection matrix. Hence, we use the learned metrics to project data into the new learned spaces. Figures 1.b, 1.d, and 1.f, respectively, show the results of the projected data which clearly illustrate the discrimination power of LMDL. Besides, the remarkable growth of the leave-one-out NN's accuracies is another evidence proving the power of discrimination of LMDL. Figure 2 shows the potential ability of kLMDL on a highly nonlinear dataset, e.g. a two-class artificial dataset consisting of 200 instances drawn from two concentric circles. As the figure shows, kLMDL achieved a much better discrimination than LMDL.

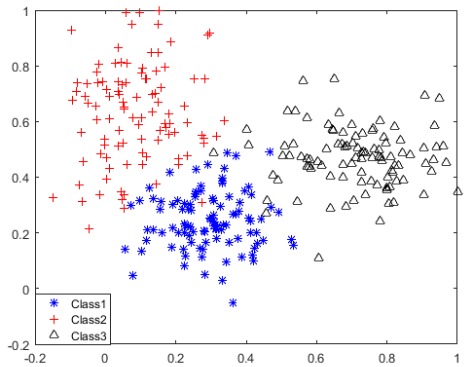

*Leave-One-Out_Accuracy_NN = 94.95*

**(a)**

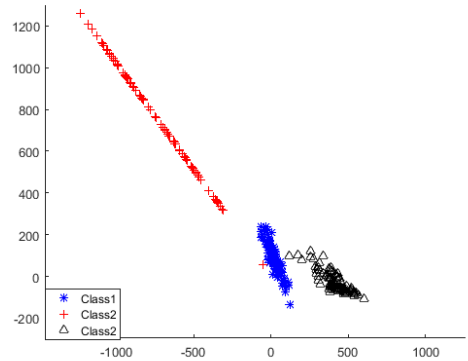

*Leave-One-Out_Accuracy_NN = 99.66*

**(b)**

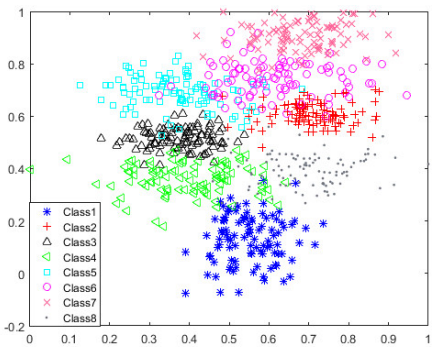

*Leave-One-Out_Accuracy_NN = 81.31*

**(c)**

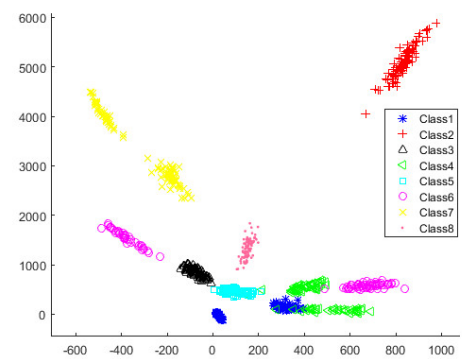

*Leave-One-Out_Accuracy_NN = 88.53*

**(d)**

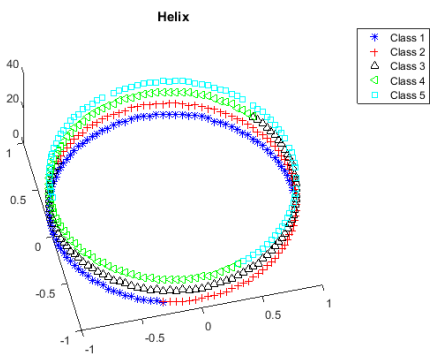

*Leave-One-Out_Accuracy_NN = 93.44*

**(e)**

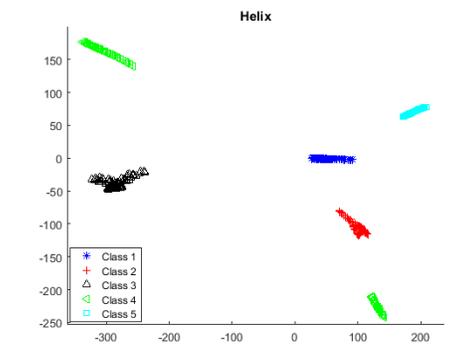

*Leave-One-Out_Accuracy_NN = 100*

**(f)**

**Figure 1.** 2D transformation of three artificial datasets achieved by LMDL. The 1NN accuracies of both dataset are also reported to validate the separation ability of LMDL.

In the next experiment, we will further explore the capability of LMDL and kLMDL from the dimension reduction point of view, i.e. learning low rank metrics. We have chosen five UCI datasets to be projected into 2D space using our proposed methods. The achieved results are shown in Figure 3. As the figure shows, the class-data separation is dramatically acceptable. It asserts that both methods are eligible for tasks of dimensionality reduction. Based on the results, we can see that the overall discriminative capacity of kLMDL is higher than LMDL. For example, in case of Australian dataset, kLMDL yields a linearly separable transformation while LMDL fails to make such a separation.

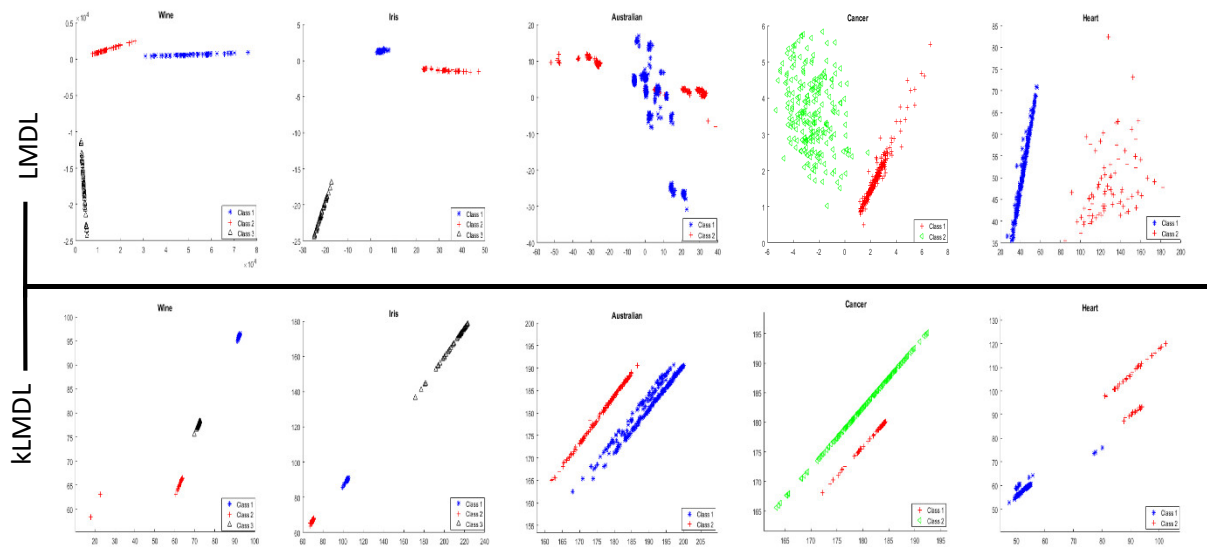

**Figure 3.** 2D visualization of some UCI datasets produced by LMDL and kLMDL. The first row shows the results of LMDL and the second row shows the same by kLMDL.

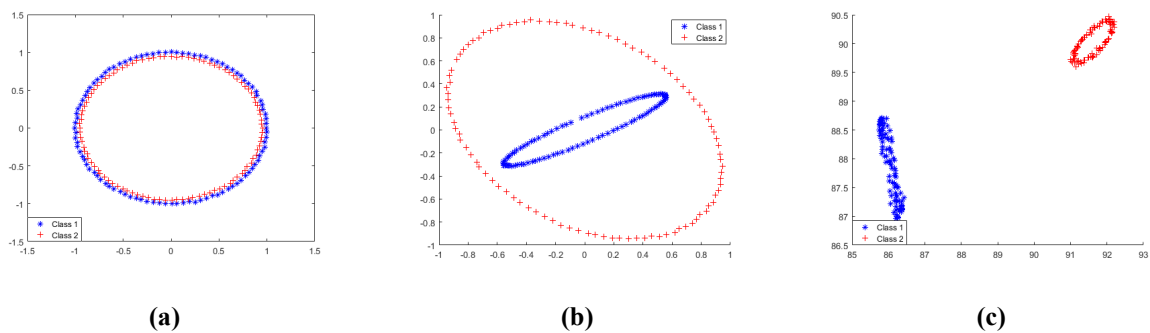

**Figure 2.** 2D transformation of a nonlinear dataset produced by the proposed methods. (a) Original space. (b) The transformed space by LMDL. (c) The transformed space by kLMDL using an RBF kernel.

In the second experiment, we apply LMDL and kLMDL on several UCI datasets with various numbers of samples, classes and dimensions. Table 1 provides a brief summary of these datasets. In the case of categorical features, we replaced each one of them by $b$ binary features where $b$ is the number of different values allowed for that features. Table 2 shows the average test error rate followed by the standard deviation (obtained by 5-time-10-fold cross validation). For some methods, the standard deviations are not reported because these results are borrowed from the authors. According to the results reported in Table 2, both LMDL and kLMDL show a superior performance compared to the rest of methods. The results of the Friedman test confirm this assertion. In order to have more reliable comparisons, the results of the Finner's post hoc procedure [30] are reported in tables 3 and 4, respectively, for LMDL and kLMDL. The Finner test is a post hoc procedure to determine whether a hypothesis of mean comparison could be rejected at a specified significance level of α. As the results show, both LMDL and kLMDL strongly reject other hypothesizes except in case of PLML against LMDL. However, the difference of p-value is not significant in this case.

**Table 1.** A brief summary of the selected UCI datasets.

| Dataset | #Samples | #Dimensions | #Classes | Dataset | #Samples | #Dimensions | #Classes |
|---|---|---|---|---|---|---|---|
| Australian | 690 | 42 | 2 | Iris | 150 | 4 | 3 |
| Balance | 625 | 4 | 3 | Letter | 20000 | 16 | 26 |
| Cancer | 685 | 9 | 2 | Liver | 347 | 6 | 2 |
| Diabetes | 768 | 8 | 2 | Seeds | 210 | 7 | 3 |
| German | 1000 | 24 | 2 | Sonar | 208 | 60 | 2 |
| Glass | 214 | 9 | 6 | Vehicle | 846 | 18 | 4 |
| Heart | 270 | 13 | 2 | Wine | 178 | 13 | 3 |
| Ionosphere | 351 | 34 | 2 | Yeast | 1484 | 8 | 10 |

**Table 2.** The average test error followed by standard deviation. Friedman statistic: distributed according to chi-square with 11 degrees of freedom: 51.43891402714929. P-value computed by Friedman Test: 3.444900729121514E-7.

| Dataset | 1NN | LPD | LMNN | LDDM | GLML | PLML | MLEV | DMLMJ [20] | KDMLMJ [20] | LMDL (Proposed) | kLMDL (Proposed) |
|---|---|---|---|---|---|---|---|---|---|---|---|
| Australian | 30.32±0.65 | 13.9 | 20.11±0.47 | 21.13±1.2 | 19.5±0.8 | 19.8±1.0 | 21.3±0.6 | 19.37±0.61 | 20.26±0.79 | 13.7±1.22 | 13.5±1.34 |
| Balance | 30.59±1.62 | 16.3 | **9.09±0.65** | 9.32±0.87 | 11.2±0.5 | 8.2±0.2 | 10.3±0.76 | 22.66±0.55 | 2.92±0.78 | 9.21±1.13 | 9.07±1.31 |
| Cancer | 4.53±0.3 | 3.4 | 3.38±1.5 | 3.18±1.2 | 3.6±0.2 | 4.9±0.7 | 3.7±1.0 | 4.22±0.18 | 4.52±0.57 | 3.22±1.5 | 3.18±1.4 |
| Diabetes | 29.32±0.75 | 26.0 | 25.86±1.07 | 26.34±2.14 | 25.12±1.09 | 26.4±0.8 | 26.7±1.3 | 26.16 | 26.78±1.2 | 25.5±1.6 | 26.33±1.12 |
| German | 32.65±0.66 | 26.0 | 31.2±0.66 | 29.76±2.4 | 30.32±1.3 | 30.43±1.3 | 33.7±2.3 | 33.38±0.73 | 31.44±1.16 | 26.2±1.32 | 26.6±1.56 |
| Glass | 27.86±1.68 | 28.0 | 34.30±1.87 | 27.62±7.03 | 31.94±1.1 | 32.18±2.3 | 31.4±7.2 | 31.21±1.08 | 32.91±2.64 | 26.33±1.87 | 27.1±2.1 |
| Heart | 24.25±1.41 | 18.6 | 19.56±0.99 | 19.45±0.76 | 19.33±0.45 | 20.11±0.8 | 20.12±2.6 | 26.7±1.17 | 23.5±1.6 | 18.5±1.42 | 18.4±1.76 |
| Ionosphere | 13.59±0.77 | 9.51±1.97 | 9.86±1.34 | 10.86±2.63 | 9.9±1.25 | 10.23±2.1 | 10.3±1.5 | 13.55±0.42 | 5.36±0.77 | 8.76±1.82 | 8.3±1.33 |
| Iris | 4.20±0.64 | 3.93±0.79 | 4.23±1.8 | 3.33±4.71 | 3.65±1.6 | 4.11±0.76 | 4.44±2.34 | 4.13±0.52 | 4.13±1.39 | 3.12±1.34 | 2.98±1.18 |
| Letter | 4.35±0.02 | **3.5** | 3.92±2.5 | 3.86±3.45 | 6.14±0.87 | 2.78±7.0 | 4.1±1.9 | 2.5 | 2.6±1.1 | 2.7±1.6 | 2.9±1.21 |
| Liver | 37.57±1.28 | 33.3 | 35.74±1.7 | 36.11±2.7 | 34.83±2.1 | 34.33±1.6 | 36.2±1.8 | 40.03±1.02 | 39.43±1.8 | 33.1±1.9 | 32.6±2.12 |
| Seeds | **6.23±0.99** | 10.33±2.80 | 6.67±3.5 | 7.23±1.0 | 6.82±2.34 | 7.17±2.33 | 7.43±2.1 | 8.71±0.67 | 11.57±1.7 | 8.44±1.78 | 7.32±1.3 |
| Sonar | 13.53±1.07 | 16.24±2.76 | 12.77±1.76 | 12.9±0.7 | 11.9±1.41 | 12.23±1.4 | 15.4±4.82 | 13.88±1.21 | 13.65±1.7 | 12.3±1.57 | 12.1±1.4 |
| Vehicle | 30.51±0.67 | 27.4 | 30.16±1.02 | 31.17±1.48 | 23.4±0.8 | 18.7±0.5 | 31.7±1.8 | 24.11±0.61 | 30.66±0.78 | 21.3±2.32 | 20.89±2.2 |
| Wine | 2.64±2.81 | 5.0 | 2.42±2.12 | 3.53±4.76 | 3.9±1.1 | 2.5±1.1 | 3.4±1.6 | 2.24±0.48 | 2.68±0.86 | 2.33±1.64 | 2.1±0.97 |
| Yeast | 47.91±0.52 | 44.87±3.24 | 46.44±0.34 | 46.32±1.14 | 46.43±1.3 | 45.98±2.1 | 47.1±3.2 | 47.19±0.71 | 47.65±0.71 | 44.7±2.62 | 44.6±2.45 |
| **Average rank** | 8.62 | 5.56 | 6.0 | 6.03 | 5.43 | 5.43 | 8.68 | 7.34 | 7.53 | 2.93 | 2.4 |

To examine the quality of the proposed methods in high-dimensional spaces, we perform another experiment on five high-dimensional datasets. Table 5 provides a summary of these datasets and compares the results in terms of errors. In the case of MNIST, Isolet, and 20NEWS [31] datasets, the training and test sets are defined. For the rest of datasets, we use 10-fold CV to obtain the results. In the case of 20NEWS, we remove each document headers as well as the stopwords.

Besides, we remove those words which appear in less than 100 documents or more than 1,500 documents. After that, we randomly select 100 documents for each category, resulting in 2000 documents. From the results, it can be seen that both LMDL and kLMDL achieve satisfactory results which are higher or comparable to the competing methods. In the case of Isolet and MNIST, the proposed methods are better and it is not significantly worse than the other cases. From the results, we can see that GLML fails to learn appropriate metrics on all datasets because its fundamental generative assumption is too restrictive and is not often valid. In contrast, our proposed methods hold no restrictive assumptions. Hence, they can be applicable in a variety of domains.

**Table 3.** Results of Finner's post hoc procedure for α=0.05, (LMDL VS other methods). Finner's procedure rejects those hypotheses that have a p-value ≤ 0.0455.

| Method | 1NN | LPD | LMNN | LDDM | GLML | PLML | MLEV | DMLMJ [20] | kDMLMJ [20] |
|---|---|---|---|---|---|---|---|---|---|
| LMDL (p-value) | 0.005 | 0.039 | 0.028 | 0.033 | 0.044 | 0.05 | 0.022 | 0.011 | 0.016 |

**Table 4.** Results of Finner's post hoc procedure for α=0.05, (kLMDL VS other methods). Finner's procedure rejects those hypotheses that have a p-value ≤ 0.0455.

| Method | 1NN | LPD | LMNN | LDDM | GLML | PLML | MLEV | DMLMJ [20] | kDMLMJ [20] |
|---|---|---|---|---|---|---|---|---|---|
| kLMDL (p-value) | 0.011 | 0.044 | 0.033 | 0.028 | 0.05 | 0.039 | 0.005 | 0.022 | 0.017 |

**Table 5.** A brief summary of three handwritten datasets along with their classification results in terms of average errors.

| Dataset | #Samples | #Dim. | #Classes | 1NN | LMNN | PLML | GLML | MLEV | LMDL | kLMDL |
|---|---|---|---|---|---|---|---|---|---|---|
| Isolet | 7797 | 617 | 26 | 8.79 | 4.49 | 4.75 | 15.97 | 15.58 | **4.3** | 4.4 |
| MNIST | 60K | 784 | 10 | 2.87 | 2.28 | 2.54 | 15.98 | 4.37 | 2.2 | **2.1** |
| Semeion | 1437 | 256 | 10 | 8.54 | **6.09** | 7.66 | 8.34 | 7.43 | 6.8 | 6.4 |
| USPS | 7291 | 256 | 10 | 5.08 | 5.38 | 6.73 | **3.75** | 6.7 | 4.7 | 5.1 |
| 20NEWS | 2000 | >5000 | 20 | 29.7 | 22.34 | 21.2 | 27.65 | 21.3 | **20.78** | 21.13 |

**Conclusion**

This paper presents a Local Mahalanobis Distance Learning (LMDL) to enhance the performance of the kNN algorithm under which the similarity of local similar points is enlarged and that of local dissimilar points is reduced as much as possible. LMDL considers the neighborhood influence and learns multiple distance metrics for a reduced set of samples, denoted as prototypes. Each prototype has its own Mahalanobis metric trying to increase local discrimination as much as possible. We use an objective function closely related to the nearest neighbor error rate in order to adjust the prototypes' metrics as well as their positions. Furthermore, the kernelized version of the proposed method is also developed to handle non-linear datasets. We have performed a variety of experiments on both synthetic and real-world datasets and the results demonstrate that the proposed method performs competitively compared with other related state-of-the-art distance metric learning methods.

**Reference**

[1] F. Wang and J. Sun, "Survey on distance metric learning and dimensionality reduction in data mining," *Data Min. Knowl. Discov.*, vol. 29, no. 2, pp. 534–564, 2015.

[2] B. Kulis, "Metric learning: A survey," *Found. Trends® Mach. Learn.*, vol. 5, no. 4, pp. 287–364, 2013.

[3] A. Bellet, A. Habrard, and M. Sebban, "A survey on metric learning for feature vectors and structured data," *arXiv Prepr. arXiv1306.6709*, 2013.

[4] R. Paredes and E. Vidal, "Learning weighted metrics to minimize nearest-neighbor classification error," *Pattern Anal. Mach. Intell. IEEE Trans.*, vol. 28, no. 7, pp. 1100–1110, 2006.

[5] M. Villegas and R. Paredes, "Simultaneous learning of a discriminative projection and prototypes for nearest-neighbor classification," in *Computer Vision and Pattern Recognition, 2008. CVPR 2008. IEEE Conference on*, 2008, pp. 1–8.

[6] Z. Hajizadeh, M. Taheri, and M. Z. Jahromi, "Nearest neighbor classification with locally weighted distance for imbalanced data," *Int. J. Comput. Commun. Eng.*, vol. 3, no. 2, p. 81, 2014.

[7] R. Paredes and E. Vidal, "Learning prototypes and distances: A prototype reduction technique based on nearest neighbor error minimization," *Pattern Recognit.*, vol. 39, no. 2, pp. 180–188, 2006.

[8] M. Dialameh and M. Z. Jahromi, "Dynamic feature weighting for imbalanced data sets," in *Signal Processing and Intelligent Systems Conference (SPIS), 2015*, 2015, pp. 31–36.

[9] M. Dialameh and M. Z. Jahromi, "A general feature-weighting function for classification problems," *Expert Syst. Appl.*, vol. 72, pp. 177–188, 2017.

[10] E. P. Xing, M. I. Jordan, S. J. Russell, and A. Y. Ng, "Distance metric learning with application to clustering with side-information," in *Advances in neural information processing systems*, 2003, pp. 521–528.

[11] J. Goldberger, G. E. Hinton, S. T. Roweis, and R. R. Salakhutdinov, "Neighbourhood components analysis," in *Advances in neural information processing systems*, 2005, pp. 513–520.

[12] A. Globerson and S. T. Roweis, "Metric learning by collapsing classes," in *Advances in neural information processing systems*, 2006, pp. 451–458.

[13] K. Q. Weinberger and L. K. Saul, "Distance metric learning for large margin nearest neighbor classification," *J. Mach. Learn. Res.*, vol. 10, no. Feb, pp. 207–244, 2009.

[14] C. Domeniconi, D. Gunopulos, and J. Peng, "Large margin nearest neighbor classifiers," *IEEE Trans. neural networks*, vol. 16, no. 4, pp. 899–909, 2005.

[15] L. Yang, R. Jin, R. Sukthankar, and Y. Liu, "An efficient algorithm for local distance metric learning," in *AAAI*, 2006, vol. 2, pp. 543–548.

[16] E. Fetaya and S. Ullman, "Learning local invariant mahalanobis distances," in *International Conference on Machine Learning*, 2015, pp. 162–168.

[17] W. Zuo *et al.*, "Distance Metric Learning via Iterated Support Vector Machines," *IEEE Trans. Image Process.*, vol. 26, no. 10, pp. 4937–4950, 2017.

[18] K. Song, F. Nie, J. Han, and X. Li, "Parameter Free Large Margin Nearest Neighbor for Distance Metric Learning.," in *AAAI*, 2017, pp. 2555–2561.

[19] L. Torresani and K. Lee, "Large margin component analysis," in *Advances in neural information processing systems*, 2007, pp. 1385–1392.

[20] B. Nguyen, C. Morell, and B. De Baets, "Supervised distance metric learning through

maximization of the Jeffrey divergence," *Pattern Recognit.*, vol. 64, pp. 215–225, 2017.

[21] M. Harandi, M. Salzmann, and R. Hartley, "Joint dimensionality reduction and metric learning: A geometric take," in *International Conference on Machine Learning (ICML)*, 2017, no. EPFL-CONF-229290.

[22] J. Wang, A. Kalousis, and A. Woznica, "Parametric local metric learning for nearest neighbor classification," in *Advances in Neural Information Processing Systems*, 2012, pp. 1601–1609.

[23] Y.-K. Noh, B.-T. Zhang, and D. D. Lee, "Generative Local Metric Learning for Nearest Neighbor Classification," *IEEE Trans. Pattern Anal. Mach. Intell.*, 2017.

[24] Y. Shi, A. Bellet, and F. Sha, "Sparse Compositional Metric Learning.," in *AAAI*, 2014, pp. 2078–2084.

[25] Y. Mu, W. Ding, and D. Tao, "Local discriminative distance metrics ensemble learning," *Pattern Recognit.*, vol. 46, no. 8, pp. 2337–2349, 2013.

[26] D. Li and Y. Tian, "Global and local metric learning via eigenvectors," *Knowledge-Based Syst.*, vol. 116, pp. 152–162, 2017.

[27] M. D. Zeiler, "ADADELTA: an adaptive learning rate method," *arXiv Prepr. arXiv1212.5701*, 2012.

[28] N. Cristianini and J. Shawe-Taylor, *An introduction to support vector machines and other kernel-based learning methods*. Cambridge university press, 2000.

[29] C. K. I. Williams and M. Seeger, "Using the Nyström method to speed up kernel machines," in *Advances in neural information processing systems*, 2001, pp. 682–688.

[30] S. García, A. Fernández, J. Luengo, and F. Herrera, "Advanced nonparametric tests for multiple comparisons in the design of experiments in computational intelligence and data mining: Experimental analysis of power," *Inf. Sci. (Ny).*, vol. 180, no. 10, pp. 2044–2064, 2010.

[31] Y. Yang, "An evaluation of statistical approaches to text categorization," *Inf. Retr. Boston.*, vol. 1, no. 1–2, pp. 69–90, 1999.